\documentclass{article}
\usepackage{spconf,amsmath,epsfig}

\usepackage{caption} 
\begin{document}\sloppy

\def\x{{\mathbf x}}
\def\L{{\cal L}}

\title{Add Pixel-level Supervision for Free: \\Weakly Supervised Localization using Background Images}
%
\name{Ziyi Kou, Wentian Zhao, Guofeng Cui, ShaojieWang}
\address{University of Rochester, Rochester 14620, United States\\\{zkou2,gcui2,swang115,wzhao14\}@ur.rochester.edu}

\maketitle

\begin{abstract}
Weakly  Supervised  Object  Localization  (WSOL)  methods usually rely on fully convolutional networks in order to obtain class activation maps(CAMs) of targeted labels.  However, these networks always highlight the most discriminative parts to perform the task, the located areas are much smaller than entire targeted objects. In this work, we propose a novel end-to-end model to enlarge CAMs generated from classification models, which can localize targeted objects more precisely. In detail, we add an additional module in traditional classification networks to extract foreground object proposals from images without classifying them into specific categories. Then we set these normalized regions as unrestricted pixel-level mask supervision for the following classification task.  We collect a set of images defined as Background Image Set from the Internet. The number of them is  much  smaller than the targeted dataset but surprisingly well supports the method to extract foreground regions from different pictures. The region extracted is independent from classification task, where the extracted region in each image covers almost entire object rather than just a significant part. Therefore, these regions can serve as masks to supervise the response map generated from classification models to become larger and more precise. The method achieves state-of-the-art results on CUB-200-2011 in terms of Top-1 and Top-5 localization error while has a competitive result on ILSVRC2016 compared with other approaches.
\end{abstract}

\section{Introduction}
Weakly Supervised Object Localization (WSOL) is one of the most important tasks in Computer Vision. It aims at localizing objects with only class-level annotations, which is challenging in that deep neural networks have access to only the classification information that tells the existence of the target object, but not the guidance of the locations of objects in an image.

\begin{figure}
    \centering
    \includegraphics[width = .45\textwidth]{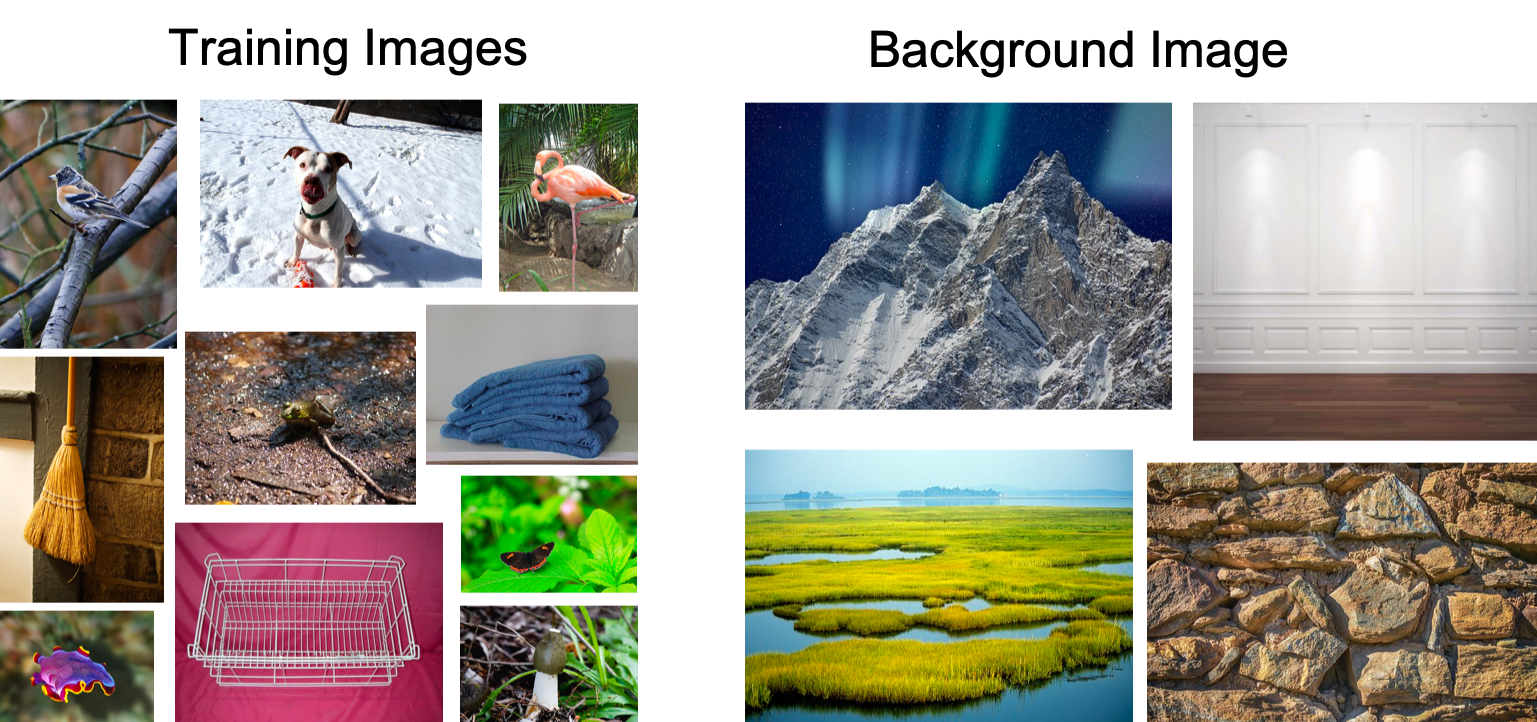}
    \caption{The images in left side are targeted ones we would like to classify. And images in right side , with only a small number, are collected from Internet which have no targeted classes but share similar background factors with the former ones.}
    \label{fig:b_f}
\end{figure}

Current WSOL methods usually locate and classify targeted objects by using convolutional classification networks. For examples, Zhou et al.~\cite{Zhou_2016_CVPR} modified pretrained classification networks to generate class activation maps (CAMs) for localization. However, CAM can only locates the most discriminative part of targeted objects, which covers only a small region of whole interests. To overcome the limitation, Wei et al.~\cite{Wei_2017_CVPR} proposed an adversarial erasing (AE) approach to discover more object parts by training several additional classification networks. Similarly, Zhang et al.~\cite{2018arXiv180406962Z} modified ~\cite{Zhou_2016_CVPR} to make the generation of CAMs end-to-end and also proposed an adversarial complementary learning approach. Moreover, Zhang et al.~\cite{2018arXiv180708902Z} proposes a self-produced model, exploring the correlations among pixels, which achieves state-of-the-art. However, the performance of all these localization methods are still mainly restricted by the classification task, which prevents the localization regions from being extended to reasonable neighbor parts.

\begin{figure*}[ht]
    \centering
    \includegraphics[width = .8\textwidth, height = .4\textwidth]{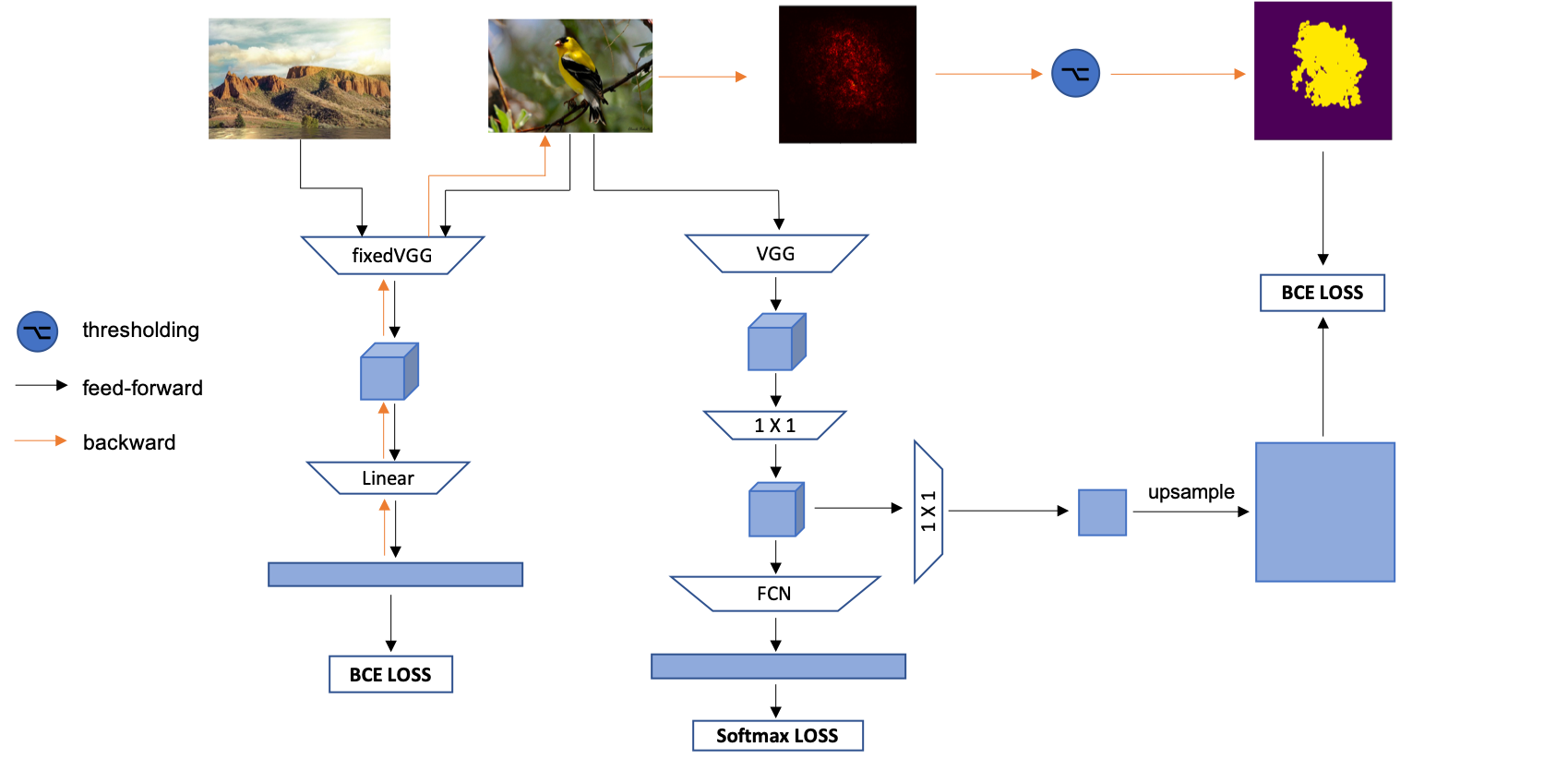}
    \caption{An illustration of our proposed model. Background Image and targeted training images are put into BC model with same probability. And the same training images are also put into CT model to be classified. The gradient map produced by BC model can be another pixel-level supervision for CAM produced by CT model. The 3-D dimensional cubes are 3-D feature maps and $1\times 1$ layers are convolution layers with $1\times 1$ kernel.}
    \label{fig:model}
\end{figure*}

As mentioned above, most fully convolutional classification networks can preserve localization information while dealing with classification tasks. Besides, some pretrained classification networks, without further finetuning on specific datasets, can also locate rough regions of objects in an image even if these objects does not appear in training or testing procedure before. Based on this, we can simply feed-forward an image and back propagate the feature map in the top layers, generating gradients on the original input. Then the foreground objects are highlighted by the high values on gradients and then can be splited from the background. However, the located regions are always larger than original objects and accompany with unrelated parts. Therefore, if we can remove these noise and regularize the corresponding ones to a more precise region, it is capable enough to be a kind of supervision mask for weakly localization tasks.

Inspired by the above motivation, in this paper, we propose a novel object extraction approach for extracting clear regions of targeted objects roughly from whole images before classification. And then we use them as pixel-level supervision to regulate CAMs in classification networks. The key idea of this extraction approach is to collect a few images, which we call Background Images, from Internet and then train a binary classification model. Since the number of images we need to collect are very small and the additional training procedure is fast, we can regard it mostly as a free obtain. After a quick training, the model not only has a high performance on binary classification, but also obtains the ability to roughly locate regions of targeted objects that are more integral and regular than the ones only produced by pre-trained models. Moreover, we can combine all of them into an end-to-end framework where same training images are input to two different models and the gradient map of binary model can be pixel-level supervision for the later classification model. To distinguish, we name the binary classification model as BC model and the model for classifying objects in targeted datasets as CT model. The whole model can be denoted as BC-CT model. 

The training structure of the whole network is illustrated in Figure \ref{fig:model}. There are two sub-modules where the left one is BC model while the right one is CT model. When training, images in training set are firstly processed with Background Images in BC model. Then the gradient of the images are served as mask supervision for the CT model. Therefore, the CT model has two supervisions where the first one is from pixel-level mask and the second one is from classification labels.

To sum up, our main contributions are:

\begin{itemize}
    \item we propose a new method to extract relative foreground regions from background using only a few images randomly selected from Internet. The regions can be further transformed to binary masks and serve as additional supervision for weakly supervised localization tasks.
    \item The proposed method outperforms the state-of-the-art approaches on CUB-200-2011 and obtains a competitive result on ILSVRC validation set.
\end{itemize}

\section{Related work}
Currently, fully supervised approaches are able enough to get a compromising performance on localization and segmentation tasks. Sermanet et al.~\cite{SermanetEZMFL13} proposed the earliest deep network, Overfeat, for object boundaries prediction. Ren et al.~\cite{NIPS2015_5638} and Lin et al.~\cite{Lin89} further developed object detection with faster RCNN and achieved great success. Furthermore, Redmon et al.~\cite{Redmon_2016_CVPR} speeds up object detection with YOLO and achieved real-time application. However, such huge achievements are based on additional manual labelled complex annotations such as bounding boxes or pixel level masks, which are expensive and sometimes ambiguous   since  there  is  nocommon  rule  for  annotating  especially  when  it  comes  topixels around object edges

Due to inefficiency and ambiguity of fully supervised approaches, WSOL methods have become more and more popular\cite{DBLP:SinghL17}\cite{2016arXiv160300489B}\cite{Wang_2014}\cite{Kim_2017_ICCV}\cite{DongZMYM17}\cite{Dong:2017}\cite{Beach}. These methods aims to inference object locations by training a deep learning model only supervised by image-level annotations. Jie et al.~\cite{Jie_2017_CVPR} propose a self-taught learning approach to firstly extract some high-response regions and then use these regions to improve the detection performance of the model. However, this method highly relies on pre-processed algorithms to get multiple region proposals. Zhou et al.~\cite{Zhou_2016_CVPR} found that global average pooling layer has ability to locate most discriminative parts of targeted classes on the feature maps in bottom layers. To enlarge such parts, Wei et al.~\cite{Wei_2017_CVPR} propose an adversarial erasing (AE) approach to retrain same images and erase most discriminative parts for several times. Zhang et al.~\cite{2018arXiv180406962Z} proposed a similar method but operated on feature maps. These methods can enlarge some additional parts of targeted objects but they destroy the integration of images, which causes some remaining parts can never be correctly classified. Moreover, Zhang et al.~\cite{2018arXiv180708902Z} again propose a self-produced model to supervise internal feature maps with each other. It introduces large number of pixel-level masks as supervision but these masks are still restricted by classification results, which limits the size and regional accuracy of final CAMs.

\section{Binary classification model}
In this section, we discuss the definition of Background Image and build the BC model learning to extract foreground regions from the background.
\subsection{Background Image}
Background Image are a few images we collect from Internet as shown in Figure \ref{fig:b_f}. They contain no targeted objects or clear labels that classification models can distinguish. However, most of them share latent background information in texture and pixel level. For example, there are lots of images taken in the nature scene which often contain similar factors, such as shining sun, crystal clear sky and boundless green grass. These images may look different in people's eyes but actually share similar features in convolutional deep networks. Therefore we can regard these collected background images as some aggregation of various background factors, which means only a few of them are enough to handle mountains of targeted samples.

\subsection{Gradient map of targeted images}
With collected Background Images, we start to build and train binary classification model. We adopt pre-trained VGG-16~\cite{VGG16} with freezed weights as feature extractor for input images denoted as $F^{b\_vgg}(I_i,\theta^{b\_vgg})$ where $\theta^{b\_vgg}$ is the parameters. And then a fully connected layer is added to produce binary value which can be denoted as $f^l(F^{b\_vgg}(I_i,\theta^{b\_vgg}))$. For each time of training, the model randomly selects images from targeted training dataset and Background Images with same probability. After training, when several query images are input, the model sets their features of fully connected layer as gradient and backward to original images, generating gradient map that highlights full region of foreground objects. The whole inference procedure can be defined as
\begin{equation}
G=backward(grad(f^l(F^{b\_vgg}(I_i,\theta^{b\_vgg}))))
\end{equation}
Where $G$ is the gradient map that has same dimensions as original images.

To use gradient maps as mask supervision for the CT model, we need to further normalize it to binary mask in which the value is $0$ or $1$ for each pixel. We define the process as follows: Given an input image of size $W\times H$, we obtain the gradient map instructed above as $G$. Then we denote the binary mask $M\in \{0,1\}^{W\times H}$, where $M_{x,y}=0$ if the pixel at $x_{th}$ row and $y_{th}$ columns is calculated to background regions, otherwise $M_{x,y}=1$ if the corresponding pixel belongs to object regions. The mask can be calculated by
\begin{equation}
M_{x,y} =
    \begin{cases}
      0 & if ~G_{x,y}<\delta*max(G)\\
      1 & if ~G_{x,y}\geq \delta*max(G)
    \end{cases}
\end{equation}
Where $\delta$ is the thresholds set as global parameter. After that, we select the largest part containing $1$ as supervision mask.

The loss function is binary cross entropy (BCE), which can be denoted as
\begin{equation}
\ell_{binary}(y,p)=-(y log(p)+(1-y)log(1-p))
\end{equation}
where $p$ is the classification results by the BC model and $y$ is background truth label. If the input is Background Image, the label is $0$. Otherwise, the label is $1$.

\begin{figure}
    \centering
    \includegraphics[width = .4\textwidth]{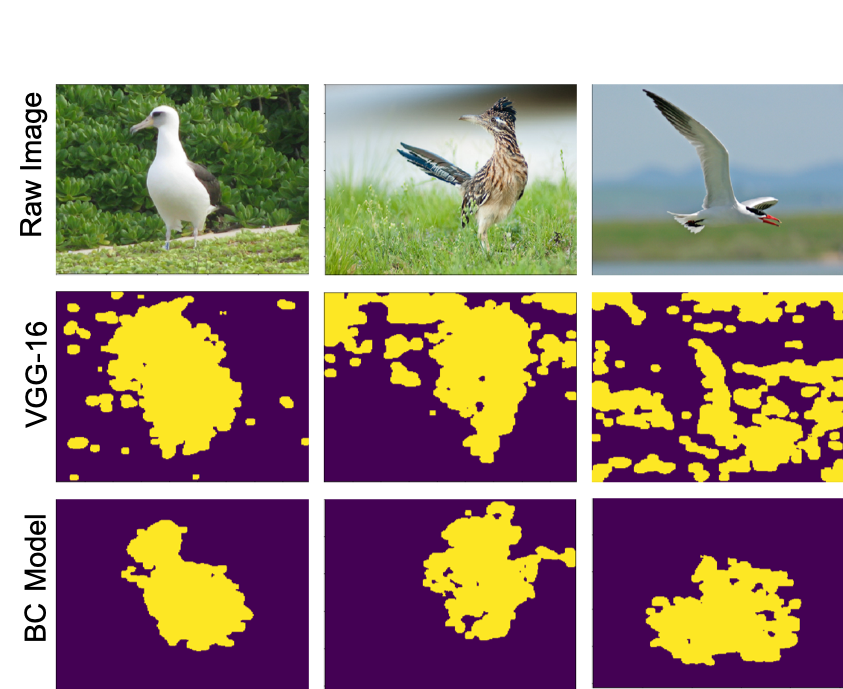}
    \caption{Comparison between VGG-16 and BC model. The first row represents orignal images. The second row contains normalized gradient maps obtained from simple VGG-16. The third row shows normalized gradient maps obtained from BC model.}
    \label{fig:3image}
\end{figure}

Some examples of processing results are shown in Figure \ref{fig:3image}. The first row is original images selected from CUB-200-2011~\cite{WahCUB_200_2011} test set. The second row contains normalized gradient maps which is obtained by backwarding only from pre-trained VGG-16 without using Background Image. The highlight areas roughly cover foreground objects but usually contain additional unrelated parts and are also sometimes irregular. The third row indicates the refined parts processed by our BC model. We can observe the model removes redundancy and re-organize reasonable parts on gradient maps, which means the Background Images we collect indeed filters unrelated background regions and re-regularize correct parts.

\section{Classification model}
The BC model provides pixel-level supervision for CT model. The latter one aims to classify different types of objects and generate CAMs for corresponding classes. As mentioned above, the original CAMs are relatively small compared with the whole targeted objects, we need to further use pixel-level mask from BC model to enlarge and regulate them. We firstly discuss the structure of the CT model and then explore if we can make two models end-to-end.

\subsection{Model structure}
As shown in Figure \ref{fig:model}, the CT model is similar with previous baseline VGG-16 model but has one additional branch. For classification task, We adopt convolutinal layers of VGG-16 as feature extractor and then add fully convolutional layer to transform the feature maps to same channels as classified labels, which can be denoted as $f^c(F^{c\_vgg}(I_i,\theta^{c\_vgg}))$ where $F^{c\_vgg}$ is the feature extractor block and $f^c$ represents fully convolutional layer. After that, a GAP layer is added to calculate average values for each channel in previous feature map followed by a fully connected layer to get final classification result. Besides, there is another branch operating on feature maps after the top fully convolutional layer. The corresponded feature map is firstly processed by two convolutional layers and then transformed to $1$ channel after a fully convolutional layer. The 1-channel feature map is upsampled to the size as same as the original image and finally a pixel-level loss is calculated between the feature map and normalized gradient map to enlarge generated CAM.

\subsection{End-to-end framework}
Since the BC model is independent from CT model and just offers supervision mask to the later one. we can train the two models simultaneously. In the training process, the training images are input to both two models and in BC model, several random selected background images are combined with training images to be the input. When obtaining gradient maps from binary classification model, we only select the gradient maps corresponding to training images and then use these maps to supervise CAMs generated from CT model.

\subsection{Loss function}
There are two different loss in CT model. The first one is cross entropy loss to classify different input images to target labels, which can be denoted as
\begin{equation}\label{eu_eqn}
\ell_{cls}(y,p)=-\sum_{c=1}^n y_{o,c}log(p_{o,c})
\end{equation}
where $n$ is the number of classes. $p$ and $y$ represents classified result and ground truth labels respectively. Besides, there is another pixel-level loss which is calculated with gradient maps from BC model and CAM from CT model denoted as
\begin{equation}\label{eu_eqn}
\ell_{mask}(M,A)=-\sum_{y=1}^n\sum_{x=1}^m M_{x,y}log(A_{x,y})
\end{equation}
where $M$ denotes the mask produced by BC model and $A$ is upsampled CAM from CT model.

We add two loss functions together as the final loss for the CT model, which can be denoted as
\begin{equation}\label{eu_eqn}
\ell_{CT}=\ell_{cls}+\ell_{mask}
\end{equation}

\section{Experiment}
\subsection{Setup}
\textbf{Implementation details} We build our model by modifying the VGG-16 network. For BC model, we freeze the weights in convolution layers and replace the final fully connected layer with a binary classifier. For CT model, we adopt the configuration of the model in~\cite{2018arXiv180406962Z}. We replace top fully connected layer with fully convolution block which contains two convolution layers of $3\times 3$ kernel size with 1024 filters and one convolutional layer of $1\times 1$ kernel size with 1000 units (200 units for CUB-200-2011). Then a GAP layer and softmax layer is followed to make classification. Besides, we also adopt another convolution block similar with above but only has one $1\times 1$ kernel size unit in final convolutional layer which transforms the channel number in the feature map from 1000 or 200 to just 1. Then this 2-D feature map is upsampled to the size of original image size and calculates mask loss.

For preprocessing of images, we have different methods for training set and validation set. For training part, we 1) randomly crop a rectangular region from original images and resize them to $224\times 224$ square. 2) randomly flip horizontally or vertically with 0.5 probability. 3) Scale hue, saturation and brightness randomly in the range $[0.6,1.4]$ 4) Normalize RGB channels respectively. For validation and testing part, we just simply resize images and then do the normalization task.

For implementation of the model, we use PyTorch framework. The model is trained for 100 epochs on NVIDIA GeForce 1080Ti with 8GB memory. The learning rate is initially 0.0001 for VGG-16 convolution layers and 0.001 for other layers then are divided by 10 for each 20 epochs.

\textbf{Datasets and evaluation} To evaluate Top-1 and Top-5 localization error of our model, we compare our model with baseline approaches on ILSVRC 2016 and CUB-200-2011 datasets. For ILSVRC 2016, 1.2 million images of 1,000 classes are contained in \textit{training} set, and 50,000 images contained in \textit{validation} set are used to test performance of models. As for CUB-200-2011 dataset, 11,788 images of 200 categories are collected, among which 5,994 images are applied for training and 5,794 images for testing. As suggested by~\cite{Russakovsky2015}, we assume correct predicted bounding box of an image as 1) Right image label are predicted for a test image. 2) More than 50 \% overlap between predicted bounding box and ground-truth box are presented.

\subsection{Results}
We compare our proposed methods with state-of-the-art methods on ILSVRC validation set and CUB-200-2011 test set.

\begin{table}
    \centering
    \begin{tabular}{l|c|c}
    \hline \hline
    Methods & top-1 err. & top-5 err.  \\\hline
    Backprop on VGGnet & 61.12 & 51.46 \\
    Backprop on GoogLeNet\cite{Szegedy_2015_CVPR} & 61.31 & 50.55 \\
    AlexNet-GAP & 67.19 & 52.16 \\
    VGGnet-GAP & 57.20 & 45.14 \\
    GoogLeNet-GAP & 56.40 & 43.00 \\
    GoogLeNet-HaS-32 & 54.53 & - \\
    VGGney-ACoL & 54.17 & 40.57 \\
    GoogLeNet-ACoL & 53.28 & 42.58\\\hline
    SPG-plain & 53.71 & 41.84 \\
    SPG & 51.40 & 40.00\\
    BC-CT(ours) & 53.21 & 42.0\\
    \hline \hline
    \end{tabular}
    \caption{Localization error on ILSVRC validation set}
    \label{tab:comp}
\end{table}

\textbf{Localization accuracy} For localization, Table \ref{tab:comp} illustrates the localization error on ILSVRC validation set of BC-CT model (ours) and several baseline algorithms. We can see that BC-CT model achieves Top-1 localization error to be 53.21 \% and Top-5 localization error to be 42.0 \%. Though the results still have a distance with state-of-the-art approach, they are far better than several models above, which indicates our method has strong ability to make an improvement. Besides, we apply our BC-CT model on CUB-200-2011 test set, the result of which is shown in Table \ref{tab:cub}. We observe that our model outperforms the state-of-the-art approach more than 2\% on both Top-1 and Top-5 results. Such result is much better than the result on ILSVRC validation set. In our opinion, the reason is that backgrounds of images in CUB-200-2011 dataset are more monotonous. For example, the birds in the images are often photoed among forests or on the sky, which can be filtered by Background Images more completely.

\begin{figure*}[ht!]
    \centering
    \includegraphics[width = 1.\textwidth]{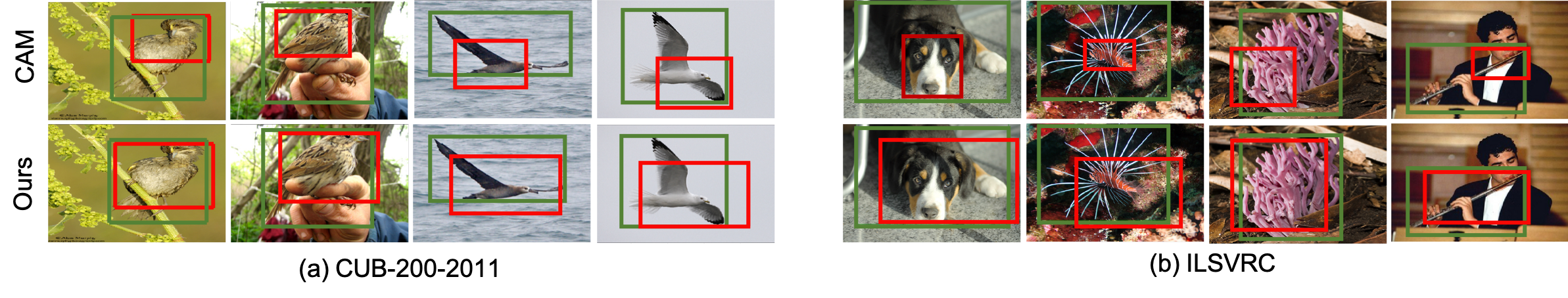}
    \caption{Comparison between CAM and BC-CT models on CUB-200-2011 and ILSVRC datasets. BC-CT models locate larger regions of objects than CAM method in both two datasets (Green bounding boxes are ground-truth, red ones are predicted).}
    \label{fig:comp}
\end{figure*}

Several results can be visualized in Figure \ref{fig:comp}. We compare our method with models without pixel-level supervision. We can see the localized region by our method is much larger than the other one. Besides, the area enlarged are usually located in correct corresponding objects, which benefits from accurate pixel-level mask as supervision. 

\begin{table}[ht]
    \centering
    \begin{tabular}{l|c|c}
    \hline \hline
    Methods & top-1 err. & top-5 err. \\\hline
    GoogLeNet-GAP & 59.00 & - \\
    ACoL & 54.08 & 43.49 \\\hline
    SPG-plain & 56.33 & 46.47 \\
    SPG & 53.36 & 42.28\\
    BC-CT (ours) & \textbf{51.21} & \textbf{40.70}\\
    BC* (ours)& \textbf{16.95} & -\\
    \hline \hline
    \end{tabular}
    \caption{Localization error on CUB-200-2011 test set}
    \label{tab:cub}
\end{table}

\textbf{Ability of pixel-level mask} ~Since the CAM of CT model is supervised by BC model, it is important for gradient map generated from the later one to cover as much parts of target object as possible to be a supervision. Therefore, we would like to see how much regions the targeted objects can be covered by mask . If the IoU of light area in gradient map and bounding box is higher than 0.5 or the former one is just larger enough to cover the whole targeted object, we set such a mask as a valid mask. The result can be seen in Table \ref{tab:cub} as BC* where the localization error is just 16.95 \%, which indicates that it's enough for gradient map to be a mask supervision for original CAM. Obviously, if we can continue leveraging the accuracy of the mask, the final localization result can be better.

\textbf{Different threshold} ~In our proposed method, the BC model provides pixel-level mask for the CT model, there is a threshold $\delta$ determining the valid range in the mask and also in the final CAM from classification model. To test its influence on localization error, we set $\delta \in \{0.7,0.75,0.8,0.85,0.9\}$ shown in in Table \ref{tab:thre}. We can see the best Top-1 error on CUB-200-2011 if $\delta=0.8$ and $\delta=0.85$ performs best on Top-1 and Top-5 error. Besides them, the performance of the model becomes worse when $\delta$ is larger or smaller. 

\begin{table}[ht]
    \centering
    \begin{tabular}{c|c|c}
    \hline \hline
    threshold & top-1 err. & top-5 err. \\\hline 
    0.7 &53.62 &43.17\\
    0.75 &53.29 &42.06 \\
    0.8 &\textbf{51.21} &41.49 \\
    0.85 &52.87 &\textbf{40.70} \\
    0.9 &52.60 &43.25 \\
    \hline \hline
    \end{tabular}
    \caption{Localization error on CUB-200-2011 with different threshold}
    \label{tab:thre}
\end{table}

\section{Conclusion}
We firstly propose a novel method to extract foreground objects from background by using a few collected images. The obtained gradient maps are independent from classification results and able to serve as mask for next classification model. Extensive results show if targeted classified dataset have a relative pure background, our method can obtain better result than other state-of-the-art methods.

{\small
\bibliographystyle{IEEEbib}
\bibliography{icme2019template}
}

\end{document}